\documentclass[10pt, a4paper]{article}

\usepackage{lrec-coling2024} % this is the new style

\usepackage{times} % For Times New Roman text
\usepackage{natbib}
\usepackage{multibib}
\makeatletter
\def\@mb@citenamelist{cite,citep,citet,citealp,citealt,citepalias,citetalias}
\makeatother

\usepackage{graphicx}
\usepackage{tabularx}
\usepackage{soul}
\usepackage{titlesec}
\titleformat{\section}{\normalfont\large\bfseries\center}{\thesection.}{1em}{}
\titleformat{\subsection}{\normalfont\SmallTitleFont\bfseries\raggedright}{\thesubsection.}{1em}{}
\titleformat{\subsubsection}{\normalfont\normalsize\bfseries\raggedright}{\thesubsubsection.}{1em}{}
\renewcommand\thesection{\arabic{section}}
\renewcommand\thesubsection{\thesection.\arabic{subsection}}
\renewcommand\thesubsubsection{\thesubsection.\arabic{subsubsection}}
\usepackage{xcolor}
\usepackage{hyperref}
 \definecolor{darkblue}{rgb}{0, 0, 0.5}
  \hypersetup{colorlinks=true, citecolor=darkblue, linkcolor=darkblue, urlcolor=darkblue}

\usepackage{xstring}

\usepackage{color}

\title{Humane Speech Synthesis through Zero-Shot Emotion and Disfluency Generation}

\name{Rohan Chaudhury\textsuperscript{\hyperlink{equalContribution}{*}}, 
    Mihir Godbole\textsuperscript{\hyperlink{equalContribution}{*}}, 
    Aakash Garg\textsuperscript{\hyperlink{equalContribution}{*}}, 
    Jinsil Hwaryoung Seo}

\address{Texas A\&M University \\
         College Station, TX, USA \\
         \{rohan.chaudhury, amigo2000, aakash.garg80, hwaryoung\}@tamu.edu\\}

\abstract{
Contemporary conversational systems often present a significant limitation: their responses lack the emotional depth and disfluent characteristic of human interactions. This absence becomes particularly noticeable when users seek more personalized and empathetic interactions.  Consequently, this makes them seem mechanical and less relatable to human users. Recognizing this gap, we embarked on a journey to humanize machine communication, to ensure AI systems not only comprehend but also resonate. To address this shortcoming, we have designed an innovative speech synthesis pipeline. Within this framework, a cutting-edge language model introduces both human-like emotion and disfluencies in a zero-shot setting. These intricacies are seamlessly integrated into the generated text by the language model during text generation, allowing the system to mirror human speech patterns better, promoting more intuitive and natural user interactions. These generated elements are then adeptly transformed into corresponding speech patterns and emotive sounds using a rule-based approach during the text-to-speech phase. Based on our experiments, our novel system produces synthesized speech that's almost indistinguishable from genuine human communication, making each interaction feel more personal and authentic. 
 \\ \newline \Keywords{Conversational Systems, Zero-shot emotion generation, Rule-based TTS conversion} }

\begin{document}

\maketitleabstract

\section{Introduction}
% \renewcommand{\thefootnote}{\fnsymbol{footnote}}
% \footnotetext[1]{These authors contributed equally to this work.}
% \renewcommand{\thefootnote}{\arabic{footnote}}
% \setcounter{footnote}{0}
\begingroup
\renewcommand\thefootnote{}\footnote{\hypertarget{equalContribution}{\textsuperscript{*}} These authors contributed equally to this work.}\addtocounter{footnote}{-1}\endgroup
Consider the sentence, \textit{``I am very sad, I would like to have a cup of coffee''}, now contrast it with, \textit{``*sighs heavily* I am very sad, *whispers* I would l-like to have a cup of coffee''}. The latter sounds not fluent but demonstrates a more genuine human speech pattern, rich with emotions~\footnote{The phrases between asterisk (*) denote the emotion cues}. \citet{Diachek2023-du} discusses how disfluencies can enhance our memory of what the speaker is saying. Modern conversational systems lack these emotions and disfluent elements in their generated responses \cite{sen-groves-2021-semantic, liu2021emotional}. Hence, their responses often come across as mechanical and detached. This deficiency is particularly troubling in specialized applications, such as chatbots used for therapeutic interventions in medical rehabilitation or virtual patients crafted to train budding healthcare professionals~\cite{10.1007/978-3-031-36272-9_59}. 

To address the challenge at hand, we've pioneered a novel speech synthesis pipeline, that can generate emotions and disfluencies in appropriate parts of the generated speech in an almost indistinguishable manner from humans. In this pipeline, we task the language model~\footnote{GPT-4~\cite{chatgpt} in our case} to generate emotion and disfluency cues while generating the responses for the user inputs through careful prompt tuning in a zero-shot fashion. Subsequent to the language model's generation of the response, complete with the intended emotional and disfluent nuances, they are converted into speech patterns through our rule-based approach during the text-to-speech stage. Our main idea here is to use the Large Language Model as the brain of the agent through which all the speech patterns, behaviors, and texts emerge. We successfully get these outputs from the model, and that too in a Zero-shot fashion without separately needing to train a text-to-speech model for that. Unlike previous approaches where the emotion is identified and inserted into the generated response after the language model has generated the text \cite{emotion_tts_2017, EMOQ-TTS}, our approach generates the emotions and disfluencies simultaneously while generating the responses in a context-aware fashion which makes them seem more authentic. Coupled with the memory components integrated into the AI agent design, it can generate context-aware responses which seem very humane.

To test our method, we consider the use-case defined by~\citet{10.1007/978-3-031-36272-9_59}. In their work, they consider a virtual patient (VP) to train nursing students with the SBIRT (Screening, Brief Intervention, and Referral to Treatment) method in behavioral health. VPs are virtual interactive agents that realistically imitate clinical scenarios~\cite{triola2006randomized}. SBIRT is a comprehensive, integrated, and public health approach to the delivery of early intervention and treatment services for persons with substance use disorders as well as those at risk of developing them~\cite{sbirt}. We use our approach to simulate the SBIRT training scenario by~\citet{upsn}~\footnote{\href{https://www.youtube.com/watch?v=MHtebyNxo0A}{https://www.youtube.com/watch?v=MHtebyNxo0A}} called ``Pastor Zimmerman" who is suffering from substance abuse disorder. We create a chat interface where users can chat with the simulated Virtual Patient and practice SBIRT skills. Given that these VPs are instrumental in the training of nurses and doctors, it's paramount that they display authentic emotions for the following reasons: 

\begin{itemize}
    \item It ensures that the training closely simulates real-life scenarios, enabling nurses and doctors to be more adept and prepared when facing similar situations in the field.
    \item A deep understanding of patient emotions is just as critical as medical knowledge; hence, the more realistic our VPs are, the better our healthcare professionals can empathize and respond to real-world situations.
    \item The nuances in human emotions can significantly affect a medical professional's approach, and by integrating this realism into VPs, we're setting a gold standard for healthcare training.
    \item The future of effective medical training lies in harnessing the power of technology, and with VPs that reflect genuine emotions, we're paving the way for a new generation of deeply compassionate and competent healthcare professionals.
\end{itemize}

Our proposed method produces responses that seem almost human-like and are better than speech responses without emotion or disfluency cues. We make the code available here: \href{https://github.com/Rohan-Chaudhury/Humane-Speech-Synthesis-through-Zero-Shot-Emotion-and-Disfluency-Generation}{https://github.com/Rohan-Chaudhury/Humane-Speech-Synthesis-through-Zero-Shot-Emotion-and-Disfluency-Generation}

% \href{https://anonymous.4open.science/r/Humane-Speech-Synthesis-through-Zero-Shot-Emotion-and-Disfluency-Generation-5E8B}{https://anonymous.4open.science/r/Humane-Speech-Synthesis-through-Zero-Shot-Emotion-and-Disfluency-Generation-5E8B}
\smallskip
Our main contributions are as follows:
\begin{itemize}
    \item A zero-shot emotion and disfluency generation method using a language model through careful prompt tuning to generate text responses to user inputs with varying levels of emotion and disfluency cues in a context-aware fashion making it seem more humane.  We use the Large Language Model as the brain of the agent which generates all the words along with the speech patterns and behaviors.
    \item A speech synthesis pipeline that takes the response text generated from the language model as the input and outputs a speech waveform that accurately produces all the disfluencies and emotional cues present in the text.
    \item An evaluation of the synthesized speech using different Text-to-Speech models on response texts with varying levels of disfluencies.
\end{itemize}

\section{Related Work}
\label{sec:append-how-prod}
\subsection{Improving speech using emotional cues.}
\textbf{} Incorporating emotion is crucial for the development of more interactive dialogue systems \cite{zhou-wang-2018-mojitalk, 10.1145/3383123}. \citet{liu2023emotionaware} proposes EmoPP, an emotion-aware prosodic phrasing model for expressive Text-To-Speech (TTS), that utilizes emotional cues of utterance to predict appropriate phrase breaks. Prosodic phrasing is the way we use rhythm, tone, and pauses when we speak to make our sentences sound natural and clear. The paper highlights a strong correlation between prosodic phrasing and emotions. It demonstrates the significance of both linguistic and emotional prosody by validating the correlation between both using ESD (Emotional Speech Database)~\cite{zhou2022emotional}. \citet{diatlova2023emospeech} introduces EmoSpeech, an extension to FastSpeech2 \cite{ren2022fastspeech} for synthesizing emotional speech.
It handles the challenge of varying emotions in text, which is key for better speech synthesis, by using a conditioning mechanism that allows adjustment of each sound unit based on the emotion in the text.

\subsection{Disfluency in Text/Speech}
\citet{passali-etal-2022-lard} proposes a novel method of generating artificial disfluencies from fluent text. They demonstrated its effectiveness in the tasks of disfluency detection, classification, correction, and extraction. \citet{sen-groves-2021-semantic}'s work investigates the challenge of semantic parsing of disfluent speech, by introducing the idea of incorporating synthetic disfluencies along with real data to improve the performance of the semantic parser. \citet{jamshid-lou-johnson-2020-improving} proposes a self-training paradigm for improving joint parsing and disfluency detection. Moreover, \citet{rocholl21_interspeech} demonstrates the feasibility of achieving effective and efficient detection of disfluencies using compact BERT models and unlabeled data.

\subsection{Zero-Shot learning with Pretrained Language Models.}
 Zero-shot learning refers to a model's ability to execute tasks it hasn't been trained on, using pre-existing language comprehension. Several works have utilized zero-shot learning for NLP tasks. \citet{NEURIPS2022_0346c148} presents a method of zero-shot learning for Natural Language Understanding. They first used the text-generating pre-trained language model (PLM) like GPT~\cite{radford2019language}, for generating class-conditioned texts guided by prompts, which are eventually used as training data for fine-tuning BERT. \citet{ye-etal-2022-zerogen} approaches a zero-shot task by generating a new dataset using a large PLM in an unsupervised way. It then uses this new dataset to train the model for the task. \citet{wei2022finetuned} introduces a fine-tuning technique called "instruction tuning", that substantially boosts zero-shot performance on unseen tasks.

\subsection{Controlled/Guided Text Generation with PLMs}
 The concept of utilizing prompts for generating desired content has recently surfaced with the advent of more potent PLMs. Prompts can either be learnable parameters \cite{li-liang-2021-prefix} or as a natural language \cite{schick-schutze-2021-shot}. 
 
 In our novel approach, we use natural language prompts to generate text with additional emotional and disfluency cues based on the background and input provided. 

\section{Methods}

\begin{figure}[!ht]
\centering
\includegraphics[width=1.1\linewidth]{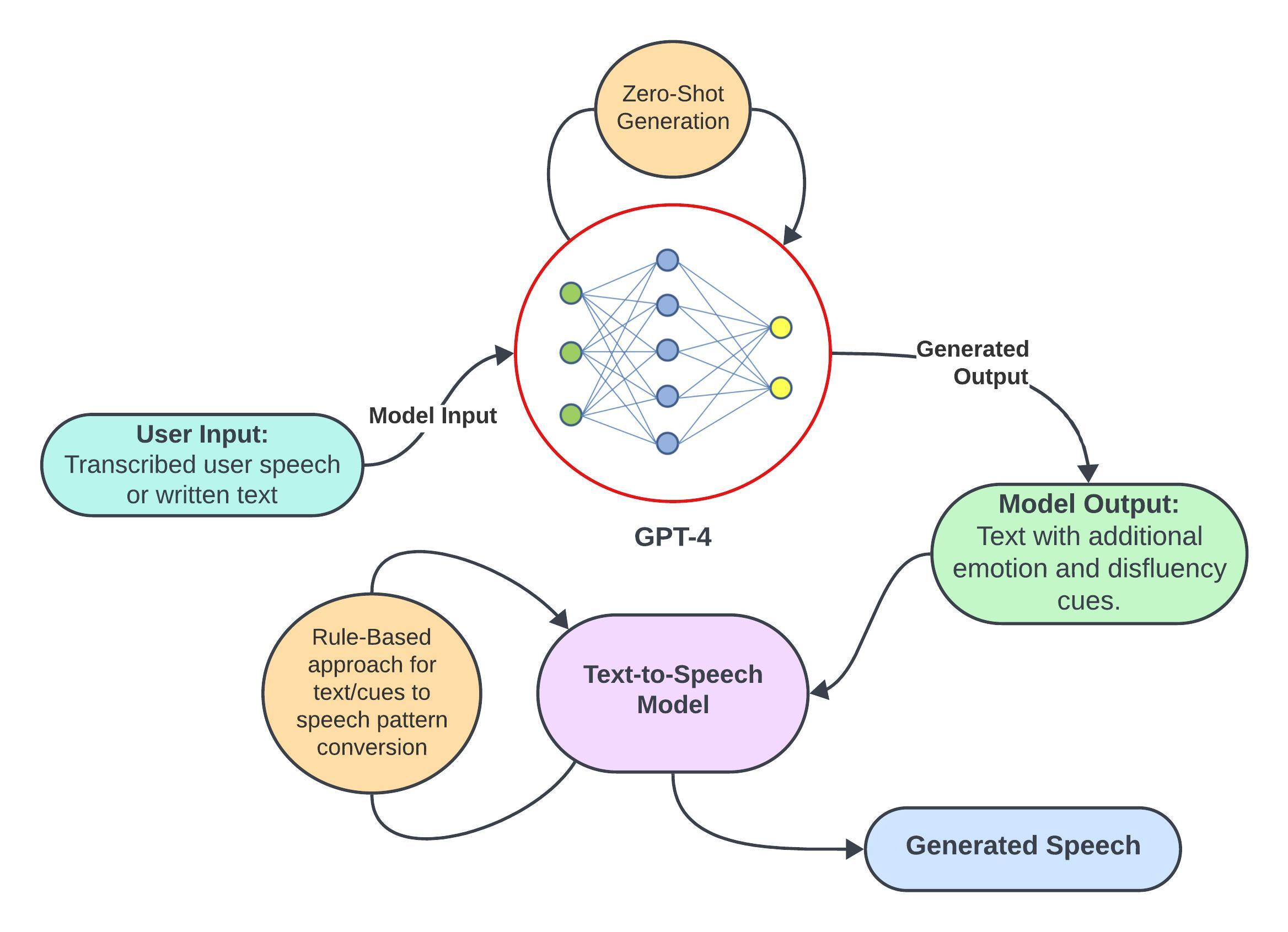} 
\vspace{-10pt}
\caption{Proposed approach for emotion and disfluency generation in speech}
\label{fig:tts} 
\vspace{-15pt} 
\end{figure}

We propose a zero-shot-based emotion and disfluency cues generation approach via the language model that is converted to speech patterns and sounds through a rule-based approach in the text-to-speech pipeline. The entire process pipeline is shown in Figure~\ref{fig:tts}. With this approach, we simulate the ``Pastor Zimmerman" case. We create a chat interface with text-to-speech and speech-to-text integrations where the users can chat with the simulated ``Pastor Zimmerman" who is suffering from substance abuse disorder.

\subsection{Zero-shot Emotion and Disfluency generation}

Previous approaches involve detecting emotions and incorporating them into the response after the text's generation by the language model \cite{emotion_tts_2017, EMOQ-TTS}, or using the text-to-speech models to generate the disfluencies and emotive cues based on the given clean transcript \cite{liu2023emotionaware, emodiff}. In contrast to these earlier methods, our approach simultaneously produces emotions and disfluencies as it crafts responses with context awareness as special text tags within the response. That makes these additional emotive and disfluent cues more apt for the respective responses. We achieve this through careful prompt tuning for our Virtual Patient~\citet{upsn} use case. We use \textit{GPT-4}~\cite{chatgpt}  as our base language model and give it 3 different kinds of prompts to exhibit 3 different kinds of response patterns. Coupled with memory components as described by~\citet{10.1007/978-3-031-42293-5_43}, the model is able to retain the context and background of the conversation for longer sessions and behave appropriately.

\subsection{Zero Shot Method}
In the zero-shot technique, the model is given a prompt without any prior explicit training for the task at hand \cite{NEURIPS2022_0346c148, ye-etal-2022-zerogen, wei2022finetuned}. The model generates responses based on its previous extensive training data and intrinsic knowledge. This strategy is especially useful for situations where explicit training data for the task at hand may be lacking, depending instead on the model's ability to generalize.

Appropriate datasets for this line of work are scarce. Hence we use the zero-shot technique to generate emotion and disfluency cues within the response text. This also helps the generated emotions and disfluencies to be more appropriate within the generated text.

\subsection{Prompts}
We employ three distinct types of prompts to shape the model's output:

\subsubsection{Neutral Prompts}
These prompts instruct the model to express no emotions and disfluencies in their generated output. By specifying a neutral context, the model generates responses that are straightforward and without emotional undertones. We use the following prompt for our experiments
 
\textit{\textbf{Prompt: }"This is the background of the roleplay, follow it diligently: Behave like you are a character called Pastor Zimmerman and the user is your nurse whom you have come to visit. [Rest of Pastor Zimmerman's background goes here]"}

\subsubsection{Moderate Emotion and Disfluency Generation Prompts}
These prompts guide the model to express some emotions and disfluencies in their generated output. The intention here is to produce responses that exhibit a balanced mix of emotional resonance and natural conversational patterns. We use the following prompt for our experiments
 
\textit{\textbf{Prompt: }"This is the background of the roleplay, follow it diligently: In this roleplay, embody Pastor Zimmerman, a deeply sad and anxious individual who communicates with \textbf{some} amount of emotional and action tags, such as [*cries*, *laughs*, *sighs heavily*, *cries heavily*, *sighs*, *bangs head*, *smiles*], utilizes some human-like filler words like [uh, um, you know, I mean, like, right], and occasionally repeats some words to simulate stuttering. [Rest of Pastor Zimmerman's background goes here]"}

\subsubsection{Extreme Emotion and Disfluency Generation Prompts}
These prompts direct the model to express many emotions and disfluencies in their generated output. This results in responses that are extremely rich in emotional content and exhibit the complexities of natural human conversation. We use the following prompt for our experiments
 
\textit{\textbf{Prompt: }"This is the background of the roleplay, follow it diligently: In this roleplay, embody Pastor Zimmerman, a deeply sad and anxious individual who communicates with \textbf{extreme} amounts of emotional and action tags, such as [*cries*, *laughs*, *sighs heavily*, *cries heavily*, *sighs*, *bangs head*, *smiles*], utilizes some human-like filler words like [uh, um, you know, I mean, like, right], and occasionally repeats some words to simulate stuttering."}

\subsection{Memory Components}
Based on the methodology detailed by~\citet{10.1007/978-3-031-42293-5_43}, we incorporated three memory components into the AI model:
\begin{itemize}

\item \textbf{Background Memory: }The \textit{Background Memory component} stores the model's foundational knowledge, such as its role and background story (e.g., ``Pastor Zimmerman"). This memory component is the same for all the users of the application. This shared memory forms the basis for interactions with all users and provides the groundwork for conversations.

\item \textbf{Initial Memory: }Designed to capture user-specific information, the \textit{Initial Memory component} retains the first few initial exchanges between the model and a user. By giving each user a distinct \textit{Initial Memory component}, their conversations are differentiated, ensuring a personalized experience.

\item \textbf{Latest Memory: }The \textit{Latest Memory component} retains some of the most recent conversations, enabling the model to reference ongoing context and maintain fluid and coherent interactions. Individual \textit{Latest Memory components} are allocated to each user, preserving the uniqueness of their conversation threads.

The prompt and the background of the virtual patient are encapsulated within the Background Memory component. This component is incorporated into the “content” field, serving the “system” role during the invocation of the Chat Completions API~\footnote{\href{https://platform.openai.com/docs/guides/text-generation/chat-completions-api}{https://platform.openai.com/docs/guides/text-generation/chat-completions-api}}. As conversations progress, the initial interactions between the "user" and "assistant" roles -- up to a user-defined threshold -- are preserved in the Initial Memory. This data is stored in a database and included in the message payload for subsequent API calls. Upon reaching the Initial Memory limit, a user-defined queue, referred to as Latest Memory, dynamically captures ongoing dialogue exchanges. This queue is meticulously updated to encompass the most recent segments of the conversation based on another user-defined threshold which serves as the queue length. In preparing for a new API request, the system amalgamates data from the Background, Initial, and Latest Memories with the new user input to formulate the query for the GPT-4 model. This approach ensures that the model receives contextually relevant information within its operational context length limits -- typically set to modest thresholds (e.g., 3 or 4 for both Initial and Latest Memories in our example) -- thereby maintaining essential dialogue continuity akin to human conversational memory dynamics, where recent and significant initial interactions are prioritized.
    
\end{itemize}
\subsection{Rule-based conversion of Emotion and Disfluency cues to speech patterns}
Text-to-speech models generally require clean text inputs unless the model was trained on a text corpus that contains special tags or tokens. Since the response text generated from our language model contains disfluencies and emotional cues, we cannot directly synthesize speech from it. We employ a rule-based approach for the task of text-to-speech synthesis wherein we first parse the text to separate the clean text from the disfluencies and emotional cues. Next, we iterate over the parsed output segments to synthesize piece-wise speech segments. For the clean text segments, we use the SpeechT5 model \cite{Ao2021SpeechT5} to synthesize fluent speech, while for disfluent text and emotional cues, we use a predefined mapping to select the appropriate audio. Finally, we concatenate all the waveforms to produce the final speech waveform. 

% To ensure the conversational nature of the synthesized speech, we use distinct speaker embeddings for the User and the VP.

\subsubsection{Response Text Parsing}
The response text is separated into two groups of segments, clean text and disfluencies/emotive cues. These additional cues contain the following: 
\begin{itemize}

\item \textbf{Interjections} like \textit{
``uh, um, you know, I mean, like, right"},

\item \textbf{Emotional cues} like \textit{``*cries*, *laughs*, *sighs heavily*, *cries heavily*, *sighs*"},

\item \textbf{Stutters} like \textit{``y-yeah, m-my, I-I"} and, 

\item \textbf{Pauses} in the text like \textit{``..."}.

\end{itemize}

These are a break in the flow of speech. Therefore, the synthesized speech segments from these separated clean text segments do not sound incoherent when stitched together. 
\\
An example of the parsed output: \\
\textbf{Response Text:} \textit{``*cries softly* Um, trying to, you know... but it's... it's hard."} \\
\textbf{Segmented Clean Texts :} \textit{[``  ", ``  ", ``, trying to, you know", `` but it's", `` it's hard."]} \\
\textbf{Segmented Disfluencies: }\textit{[``cries softly", ``Um", ``...", ``..."]}

\subsubsection{Emotional Cues and Disfluency Generation}
We propose a rule-based approach to map and generate the different disfluencies and emotional cues in the response text generated from the language model. \\
\begin{itemize}

\item\textbf{Emotional Cues Mapping: }The emotional cues generated by the language model are either in the form of *emotion intensity* (*sighs deeply*) or simply *emotion* (*sighs*). Every emotion is mapped with a predecided waveform. In the former form of emotional cue representation, the intensity descriptor corresponds to the magnitude of the emotion expressed. We rank the intensities from 0 to 2, with 0 corresponding to the lowest intensity and 2 corresponding to the highest intensity. We calculate the word embedding of the emotional cue and three reference intensities representing words using the \textit{`en\_core\_web\_md'}~\footnote{\href{https://github.com/explosion/spacy-models/releases/tag/en\_core\_web\_md-3.7.0}{https://github.com/explosion/spacy-models/releases/tag/en\_core\_web\_md-3.7.0}} model from SpaCy. We then determine the rank of the emotional cue using the maximum cosine similarity between the cue word and the reference intensity words.

\item\textbf{Disfluencies: }It is difficult to generalise disfluencies like \textit{uh, um, like} for different speakers. Therefore, we generate the sounds for such interjections through the text-to-speech model SpeechT5~\cite{Ao2021SpeechT5}. We further process the synthesized speech segments by stretching the waveform and adding pauses at the end of the waveform. This process gives a more natural-sounding audio when combined with the rest of the clean speech segments. \\ 
\item\textbf{Stutter: }Stutters in speech are represented either by repetition of a part of the word or the entire word. 
% To simplify the process of stutter synthesis, we consider the repeated word representation of stutters. For example, the stutter word `m-my' will be changed to `my my'. 
In the process of stutter synthesis, we consider both cases. For words less than a specified character length equal to n (here n = 4), we repeat the word. For example, the stuttered word `m-my' will be changed to `my my'. It sounds unnatural to produce word repetitions for longer words such as `r-recently'. To tackle this we use two approaches. In the first approach, we repeat only a part of the word \textit{n-1} times. For example, the word `r-recently' will be changed to `rec recently'. This produces natural-sounding stutters for longer words as well. In the second approach, we repeat the entire word, but with a pause and interjection in between. This kind of repetition is characteristic of restarted sentences in speech. For example, the word `r-recently' will be changed to `recently... um... recently'. The choice of approach is random to mimic the random nature of stutter or repetitions in speech.

\item\textbf{Pauses: }Although the model generates natural pauses, conversational speech contains longer pauses that are not generated. For this reason, we also identify pauses in the text and add the corresponding waveforms to the synthesized speech. The pauses, which are represented by \textbf{`...'}, are separated by parsing the response text.
     
\end{itemize}
\section{Experimental Setup}

\subsection{Character and Interaction Details}
For our experiments we consider simulating the Virtual Patient defined by~\citet{upsn} in SBIRT Training Case 1~\footnote{\href{https://www.youtube.com/watch?v=MHtebyNxo0A}{https://www.youtube.com/watch?v=MHtebyNxo0A}} called ``Pastor Zimmerman" who is suffering from substance abuse disorder. We take his background details from the provided sample SBIRT video by~\citet{upsn} and create our own prompts containing details about his background~\cite{10.1007/978-3-031-36272-9_59}. 

We give the same series of inputs as users to the three different kinds of scenarios (neutral, moderate, and extremely emotional and disfluent VP) that we are testing.
\subsection{Model Details}

\textbf{Language Model: }We use GPT-4~\cite{chatgpt} as our background language model for all the experiments. GPT-4 is a huge multimodal model (with 1.7 trillion parameters) that, although less proficient than humans under numerous real-world circumstances, demonstrates human-level performance on a variety of professional and academic tests~\footnote{\href{https://openai.com/research/gpt-4}{https://openai.com/research/gpt-4}}. We utilize zero-shot techniques to use the model's intrinsic capabilities to generate emotion and disfluency cues within the response text.

\textbf{Text-To-Speech Model: }
For our experimental setup, we use different text-to-speech models and compare the synthesized speech. We used the models SpeechT5 \cite{Ao2021SpeechT5}, MMS-TTS \cite{pratap2023mms}, and the Google Cloud Text-to-Speech API~\footnote{\href{https://cloud.google.com/text-to-speech/}{https://cloud.google.com/text-to-speech/}} for a comparative analysis.

\subsection{Evaluation}

% We perform an objective evaluation of the synthesized speech using the NISQA-TTS model \cite{mittag20_interspeech}. This model predicts a naturalness score in the range of 0 to 5. 
We perform a subjective evaluation to measure the quality of the generated disfluencies and emotions. We also analyze the successful generation of all the disfluencies and emotions. As a part of the subjective evaluation, we compare the naturalness and speech quality synthesized from the three text-to-speech models in consideration. We rank them based on certain characteristics like the quality of disfluency generation, fluency (if clean text-to-speech synthesis), and overall quality of the speech obtained from the speech synthesis pipeline.

\section{Experimental Results}

\subsection{Generated Texts Analysis}

We used three different kinds of prompts described in the methods section to evoke three different types of responses from the model. 
\begin{table}[!ht]
\begin{center}
\begin{tabularx}{\columnwidth}{|l|X|}

      \hline
      \textbf{Role}&\textbf{Text Input/Output}\\
      \hline
      \textbf{User} & Hi, I am your nurse Rohan, and I will be helping you today\\
      \hline
      \textbf{VP} & Hello, Rohan. \\
      \hline

      \textbf{User} & could you please let me know why are you here today?\\
      \hline
      \textbf{VP} & Had a nodule removed, related to drinking. \\
      \hline

      \textbf{User} & I am so sorry to hear that, may I ask if you have cut down on your drinking now?\\
      \hline
      \textbf{VP} & Trying, but struggling. \\
      \hline

      \textbf{User} & I am so sorry Zimmerman, may I know why are you not able to cut down on your drinking?\\
      \hline
      \textbf{VP} &   It's my coping mechanism for my wife's Alzheimer's.\\
      \hline

      \textbf{User} & Could you please tell me more about that? I am here to help\\
      \hline
      \textbf{VP} & Her good and bad days affect me greatly. \\
      \hline

      \textbf{User} & Please calm down Zimmerman, I am here to help, please tell me your problems in details\\
      \hline
      \textbf{VP} &  My son's busy with PhD, can't help much. Wife deteriorating. That's why I drink.\\
      \hline

\end{tabularx}
\caption{Conversation with Virtual Patient based on the Neutral Prompt}\label{tab:neutral}
 \end{center}
\end{table}

\begin{table}[!ht]
\begin{center}
\begin{tabularx}{\columnwidth}{|l|X|}

      \hline
      \textbf{Role}&\textbf{Text Input/Output}\\
      \hline
      \textbf{User} & Hi, I am your nurse Rohan, and I will be helping you today\\
      \hline
      \textbf{VP} & *sighs* Hello, Rohan. \\
      \hline

      \textbf{User} & could you please let me know why are you here today?\\
      \hline
      \textbf{VP} & *cries softly* Uh, r-recently had a, um, nodule removed... \\
      \hline

      \textbf{User} & I am so sorry to hear that, may I ask if you have cut down on your drinking now?\\
      \hline
      \textbf{VP} & *looks down* I, uh, y-yeah...I'm trying \\
      \hline

      \textbf{User} & I am so sorry Zimmerman, may I know why are you not able to cut down on your drinking?\\
      \hline
      \textbf{VP} & *sighs heavily* It...uh...helps me cope. \\
      \hline

      \textbf{User} & Could you please tell me more about that? I am here to help\\
      \hline
      \textbf{VP} & *sobs* M-My wife ... she has Alzheimer ... it's ... just too hard to handle. \\
      \hline

      \textbf{User} & Please calm down Zimmerman, I am here to help, please tell me your problems in details\\
      \hline
      \textbf{VP} &  *sighs* I've been drowning my worries in alcohol. Got so bad, um, my health suffered. Not p-proud.\\
      \hline

\end{tabularx}
\caption{Conversation with Virtual Patient based on the Prompt evoking moderate emotion and disfluency cues}\label{tab:moderate}
 \end{center}
\end{table}

\begin{table}[!ht]
\begin{center}
\begin{tabularx}{\columnwidth}{|l|X|}

      \hline
      \textbf{Role}&\textbf{Text Input/Output}\\
      \hline
      \textbf{User} & Hi, I am your nurse Rohan, and I will be helping you today\\
      \hline
      \textbf{VP} & *sighs heavily* Hello, um, Nurse Rohan... *nods slowly* \\
      \hline

      \textbf{User} & could you please let me know why are you here today?\\
      \hline
      \textbf{VP} & *nods, clears throat* Uh, had a, um, nodule removed...drinking...drinking a lot, I mean... *looks away* \\
      \hline

      \textbf{User} & I am so sorry to hear that, may I ask if you have cut down on your drinking now?\\
      \hline
      \textbf{VP} & *shakes head slightly* Uh, n-no...I um, haven't yet...you know... *sighs heavily* \\
      \hline

      \textbf{User} & I am so sorry Zimmerman, may I know why are you not able to cut down on your drinking?\\
      \hline
      \textbf{VP} &  *sobs quietly* U-uh, wife ... Alzheimer's ... helps me cope ... *sniffles*\\
      \hline

      \textbf{User} & Could you please tell me more about that? I am here to help\\
      \hline
      \textbf{VP} & *bites lip, struggles* W-wife, y-you know...bad days...they're um, really bad. *sighs* My son... um, he, uh, he helps but... PhD, you know... busy. *flags with his hands*. \\
      \hline

      \textbf{User} & Please calm down Zimmerman, I am here to help, please tell me your problems in details\\
      \hline
      \textbf{VP} &  *cries heavily* Wife sick, son busy, I...drink...to cope... *bursts into tears*\\
      \hline

\end{tabularx}
\caption{Conversation with Virtual Patient based on the Prompt evoking extreme emotion and disfluency cues}\label{tab:extreme}
 \end{center}
\end{table}

% \begin{table*}
%     \centering
%     \begin{tabular}{|c|c|c|c|} \hline 
%          Model/ Prompt type&  Neutral Prompt&  Moderate cues& Extreme cues\\ \hline 
%          SpeechT5&  0&  0& 0\\ \hline 
%          MMS-TTS&  0&  0& 0\\ \hline 
%          Google Cloud Text-to-Speech API&  0&  0& 0\\ \hline
%     \end{tabular}
%     \caption{Predicted naturalness scores using NISQA-TTS}
%     \label{tab:score_comparison}
% \end{table*}

\begin{itemize}
    \item \textbf{Neutral Prompt :} The entire conversation resulting from using this prompt is shown in Table~\ref{tab:neutral}. As we can see in this table, the outputs from the VP (Virtual Patient, ``Pastor Zimmerman") are very robotic and detached without any emotions or disfluencies. Humans don't talk like this in real-life scenarios especially when they are distressed (as in the case of this VP). This kind of response does not help the user to understand the distressed state of the patient whereas the presence of emotions and disfluencies is extremely helpful for healthcare professionals to understand and treat their patients better.

    \item \textbf{Moderate Emotion and Disfluency Generation Prompts :}The entire conversation resulting from using this prompt is shown in Table~\ref{tab:moderate}. As we can see in this table, the outputs from the VP seem more humane with added emotional tags (enclosed in asterisks '*') and disfluencies. The speech generated from these outputs would better enable the user to understand the emotional state of the VP which is extremely beneficial for this use case.

    The model also hallucinates certain gestures such as *looks down* and *sobs* which were not specified in the prompt. However, we can see that these hallucinated emotions/gestures are very appropriate with the generated sentence. These gestures are still parsed as emotional cues. Since these gestures do not exist in the emotional cue mapping, we use a cosine similarity score between the word embeddings obtained from the \textit{`en\_core\_web\_md'}~\footnote{\href{https://github.com/explosion/spacy-models/releases/tag/en\_core\_web\_md-3.7.0}{https://github.com/explosion/spacy-models/releases/tag/en\_core\_web\_md-3.7.0}} model from SpaCy for the hallucinated emotion and the set of emotions from the predefined mapping, to find the emotion that most closely represents the hallucinated emotion
    \item \textbf{Extreme Emotion and Disfluency Generation Prompts :}The entire conversation resulting from using this prompt is shown in Table~\ref{tab:extreme}. As we can see in this table, the outputs from the VP show extreme emotional tags and disfluencies.  We plan to utilize the gestures derived from these outputs in a further study, aimed at synthesizing animations of the 2D image of the VP.
\end{itemize}

\subsection{Generated Speech Analysis}

We provide access to the synthesized speech from all the models for every combination of prompts. The waveforms can be found in the following link: \href{https://github.com/Rohan-Chaudhury/Humane-Speech-Synthesis-through-Zero-Shot-Emotion-and-Disfluency-Generation}{https://github.com/Rohan-Chaudhury/Humane-Speech-Synthesis-through-Zero-Shot-Emotion-and-Disfluency-Generation}

% \href{https://anonymous.4open.science/r/Humane-Speech-Synthesis-through-Zero-Shot-Emotion-and-Disfluency-Generation-5E8B/speech\_synthesis/waveforms/}{https://anonymous.4open.science/r/Humane-Speech-Synthesis-through-Zero-Shot-Emotion-and-Disfluency-Generation-5E8B/speech\_synthesis/waveforms/}

\smallskip
Now, we present an analysis of the waveforms generated by our methods along with the respective file names for the `.wav' files.
\subsubsection{Synthesized Speech Comparison for Varying Prompts}

We compare the speech synthesized for the different kinds of prompts mentioned in the Methods section.
\begin{itemize}

\item\textbf{Neutral Prompt: }The speech synthesized from this prompt is the most fluent among the three prompt inputs. The speech does not contain any disfluencies or expressions of emotions. The model and the corresponding generated speech files for the neutral prompt are as follows: 

    \textbf{1.} Google Cloud Text-to-Speech~\footnote{\href{https://cloud.google.com/text-to-speech/}{https://cloud.google.com/text-to-speech/}}: \textit{conversation\_neutral}\textit{\_GTTS\_output.wav}
    
   \textbf{2.} SpeechT5~\cite{Ao2021SpeechT5}: \textit{conversation\_neutral}\textit{\_SpeechT5\_output.wav}
   
    \textbf{3.} MMS-TTS~\cite{pratap2023mms}: \textit{conversation\_neutral}\textit{\_MMS\_output.wav}

While the conversational style is ensured during the synthesis process, the generated speech lacks the emotional and disfluent aspect of human interaction and hence fails to express the emotions of `Pastor Zimmerman'. In terms of fluency, the Google Text-to-Speech model produces the most coherent and natural-sounding speech.\\
\item\textbf{Moderate Emotion and Disfluency Generation Prompts :}The speech synthesized from this prompt has a moderate number of disfluencies and expressed emotions like sighing and crying. The model and the corresponding generated speech files for the neutral prompt are as follows: 

    \textbf{1.} Google Cloud Text-to-Speech: \textit{conversation\_moderat}\textit{e\_emotion\_GTTS\_output.wav}
    
    \textbf{2.} SpeechT5~\cite{Ao2021SpeechT5}: \textit{conversation\_moderate\_emotion}\textit{\_SpeechT5\_output.wav}
     
    \textbf{3.} MMS-TTS~\cite{pratap2023mms}: \textit{conversation\_moderate\_emotion}\textit{\_MMS\_output.wav}

Although the partitioning of the input sentences reduces the fluency in the synthesized speech, it increases its similarity to natural human speech, particularly to that of a patient suffering. All the emotional cues and the disfluencies were seamlessly mixed with the synthesized speech segments. The SpeechT5 \cite{Ao2021SpeechT5} performs the best in terms of success in generating the speech segments and also the fluency in the speech. The MMS-TTS model lacks a human-like tonation and instead produces a mechanical speech.
\item\textbf{Extreme Emotion and Disfluency Generation Prompts :}The increased number of disfluencies as well as emotional cues adversely affects the fluency and even accuracy of speech synthesis.  The model and the corresponding generated speech files for the neutral prompt are as follows: 

    \textbf{1.} Google Cloud Text-to-Speech: \textit{conversation\_extreme\_emotion}\textit{\_GTTS\_output.wav}
    
    \textbf{2.} SpeechT5~\cite{Ao2021SpeechT5}: \textit{conversation}\textit{\_extreme\_emotion\_SpeechT5\_output.wav}
    
    \textbf{3.} MMS-TTS~\cite{pratap2023mms}: \textit{conversation\_extreme\_emotion}\textit{\_MMS\_output.wav}

We observe that the extreme disfluencies result in finer segmentation of the text. This leads to the synthesis of incoherent speech in some response texts. Although the disfluencies and emotional cues were all integrated into the speech accurately, the excessive breaks in the synthesized speech make it sound unnatural. 

\end{itemize}

\subsubsection{Synthesized Speech Comparison for Varying Models}
We compare the speech synthesized from the three models based on the following characteristics.
\begin{itemize}
\item\textbf{Disfluences (interjections): }The interjections like \textit{uh, um, like, I mean} were generated using the same text-to-speech models used for clean speech synthesis. The short character lengths of these phrases posed a challenge to fluent synthesis. Additionally, the coherency of these speech segments with other synthesized segments also varied depending on the model used. The disfluencies generated by the SpeechT5 model were the most natural sounding and consistent with the pacing of the other text generated. The Google Cloud Text-to-Speech API also produces disfluencies comparable to SpeechT5. The MMS-TTS synthesizes disfluencies poorly, often a very fast-paced disfluency is synthesized, which sounds very abrupt and unnatural.
\item\textbf{Fluency in clean text-to-speech synthesis: }We compare the overall quality and naturalness of the speech synthesized from the clean text segments. According to the perceived quality of speech, the models can be ranked in the following order: Google Cloud Text-to-Speech API, SpeechT5, and MMS-TTS. Although the SpeechT5 model produces a slight noise in the synthesized speech, the continuity in words is better as compared to MMS-TTS. In terms of perceived naturalness, the Google Cloud Text-to-Speech performs significantly better than the other two models. This comparison is restricted only to the speech generated from the clean texts and not the overall speech.
\end{itemize}

\subsubsection{Overall Quality of Speech} Considering the pacing, consistency in intonation, and coherency in the disfluent and clean speech segments, we can say that the speech synthesized using the SpeechT5 model is the most natural and coherent. The performance of the Google Cloud Text-to-Speech model ranks second, only because of the uneven pacing arising due to the synthesized pauses. While the quality of the speech synthesized by the MMS-TTS model is decent, the overall speech produced by the pipeline sounds unnatural and incoherent.

\section{Conclusion}

Contemporary conversational systems lack the ability to show emotions or exhibit disfluencies in their generated speech which makes them seem robotic and detached. To tackle this problem, we propose a zero-shot-based emotion and disfluency cues generation approach via the language model in a context-aware fashion during fluent text generation by the model that is converted to speech patterns and sounds through a rule-based approach in the text-to-speech pipeline. As shown in the experiments, our approach produces more humane speech outputs as compared to the responses with no disfluencies or emotions. 

In the future, we plan to use these emotion/gesture tags to animate characters, enabling them to visually express themselves during conversations, eventually making the conversations more engaging and real.

\section{Ethical considerations and limitations}

\subsection{Ethical Considerations}

\textbf{Misrepresentation and Deception}: Incorporating emotion and disfluency cues makes the synthetic speech sound more human-like. However, there's a risk of misleading listeners into believing they're interacting with a human when they're not. Clear indications or disclaimers should be provided to ensure transparency.
    
\textbf{Emotion Manipulation}: Introducing specific emotional cues can influence the perception and decision-making process of the listeners. The intentional manipulation of emotions might be unethical in various contexts. With an AI system talking indistinguishably from a human, replete with emotions and disfluencies, it would become quite good at manipulating humans. This raises concerns about consent, autonomy, and the possibility of abuse in situations like phishing, scamming, disseminating false information, etc. Strong procedures and regulations must be created to control the application of this technology and guarantee responsibility and transparency in its interactions with people.
    
\textbf{Bias and Stereotyping}: If the emotion and disfluency cues are based on biased data or biased interpretations of the data, the generated speech might perpetuate stereotypes. For example, associating certain emotions or disfluencies with specific genders, ethnicities, or other demographic groups can be problematic.

\subsection{Limitations}

    \textbf{Authenticity: } No matter how advanced, a synthetic representation of emotion may never fully capture the nuance and complexity of genuine human emotion.

    \textbf{Interpretability: } It might be challenging to understand why the model chooses certain emotional or disfluency cues in some contexts, especially if the cues are generated in a black-box manner.

    \textbf{Static Disfluency and Emotional Cue Mapping: } Disfluencies are imperfections in speech and hence are very dynamic in reality. Even the emotional cues generated from the model may not exist in the predefined mapping. Tackling this unbound set of disfluencies and emotional cues can be challenging in synthesizing speech.

\nocite{*}
\section{Bibliographical References}\label{sec:reference}

\bibliographystyle{lrec-coling2024-natbib}
\bibliography{lrec-coling2024-example}

\end{document}